\documentclass[10pt,letterpaper]{article}
\usepackage[top=0.85in,left=2.75in,footskip=0.75in,marginparwidth=2in]{geometry}

\usepackage[utf8]{inputenc}

\usepackage{cite}

\usepackage{algorithm2e}

\usepackage{nameref,hyperref}

\usepackage[right]{lineno}

\usepackage{microtype}
\DisableLigatures[f]{encoding = *, family = * }

\raggedright
\usepackage{changepage}

\usepackage[aboveskip=1pt,labelfont=bf,labelsep=period,singlelinecheck=off]{caption}

\makeatletter
\renewcommand{\@biblabel}[1]{\quad#1.}
\makeatother

\usepackage{lastpage,fancyhdr,graphicx}
\usepackage{epstopdf}
\fancyheadoffset[L]{2.25in}
\fancyfootoffset[L]{2.25in}

\usepackage{color}

\definecolor{Gray}{gray}{.25}

\usepackage{graphicx}
\usepackage{multirow}
\usepackage{amsmath}
\usepackage{amssymb}
\usepackage{subfig}
\usepackage{graphicx}

\usepackage{sidecap}

\usepackage{wrapfig}
\usepackage[pscoord]{eso-pic}
\usepackage[fulladjust]{marginnote}
\reversemarginpar

\begin{document}
\vspace*{0.35in}


\begin{flushleft}
{\Large
\textbf\newline{Adaptive Convolutional ELM For Concept Drift Handling \\in Online Stream Data}
}
\newline
\\
Arif Budiman,
Mohamad Ivan Fanany,
Chan Basaruddin
\\
\bigskip
\bf{1} Machine Learning and Computer Vision Laboratory \\Faculty of Computer Science, Universitas Indonesia\\
\bigskip
* arif.budiman21@ui.ac.id

\end{flushleft}

\providecommand{\keywords}[1]{\textbf{\textit{Keywords---}} #1}

\section*{Abstract}
In big data era, the data continuously generated and its distribution may keep changes overtime. These challenges in online stream of data are known as concept drift. In this paper, we proposed the Adaptive Convolutional ELM method (ACNNELM) as enhancement of Convolutional Neural Network (CNN) with a hybrid Extreme Learning Machine (ELM) model plus adaptive capability. This method is aimed for concept drift handling. We enhanced  the CNN as convolutional hiererchical features representation learner combined with Elastic ELM (E$^2$LM) as a parallel supervised classifier. We propose an Adaptive OS-ELM (AOS-ELM)  for concept drift adaptability in classifier level (named ACNNELM-1) and matrices concatenation ensembles for concept drift adaptability in ensemble level (named ACNNELM-2). Our proposed Adaptive CNNELM  is flexible that works well in classifier level and ensemble level while most current methods only proposed to work on either one of the levels. 

We verified our method in extended  MNIST data set and not MNIST data set.  We set the experiment to simulate virtual drift, real drift, and hybrid drift event and we demonstrated how our CNNELM  adaptability works.  Our proposed method works well and gives better accuracy, computation scalability, and concept drifts adaptability compared to the regular ELM and CNN. Further researches are still required to study the optimum parameters and to use more varied image data set. 
\bigskip

\noindent\keywords{deep learning, extreme learning machine, convolutional, neural network, big data, online,concept drift}


\section{Introduction}
Online data stream learning is an emerging research area that has shown its importance in big data era. The big volumes of data are continuously generated from devices, softwares, and Internet, with higher incoming rate and no time bound.  Knowledge mining over them needs special machine learning techniques to learn large volumes of data in a timely fashion. The techniques also need to be scalable for deployment in the big online real-time scenario (computationally tractable for processing large data streams) and capable to overcome uncertainty in the data representations.

The techniques, to offer an adaptive framework which adapts to any issue in which the data concepts do not follow static assumptions (known as concept drift problem \cite{Gama:2014:SCD:2597757.2523813,5156502}).
In concept drift (CD),  the input and/or output concepts has non stationary and uncertain data distribution. The uncertainties can be perceived by the increase of class overlapping or additional comprehensive features in the feature space, which makes a deterioration of classifiers.

The aim of CD handling \cite{Gama:2014:SCD:2597757.2523813} is to boost the generalization accuracy when the drift occurs. Common handling methods are based on classifier ensemble \cite{6779381}. 
Ensemble methods combined decision from each classifier members (mainly using ensemble members diversification). However, ensemble methods are difficult to manage complexities when handling many types of consecutive drifts \cite{39313598,978-3-540-22144-9,KunchevaEnsembleOverview08,Budiman_hindawi}.


One of the recent online big stream data approaches are based on Deep Learning (DL) techniques \cite{Bengio:2009:LDA:1658423.1658424}. They offer promising avenue of automated feature extraction of big data in streaming approaches.
The Deep Learning has many variants such as  a Deep Belief Network (DBN) by Hinton \cite{hinton:1}, Deep Boltzmann Machine (DBM) by Salakhutdinov and Hinton, \cite{SalHinton07}, Stacked Denoising Autoencoders (SDA) by Vincent \cite{vincent:1}, and Convolutional Neural Network (CNN) by LeCun \cite{LeCun:1}, and many others. 

The traditional machine learning methods i.e. Extreme Learning Machine (ELM), Support Vector Machine (SVM), Multi-Layer Perceptron Neural Network (MLP NN), Hidden Markov Model (HMM) may not be able to handle such big stream data directly \cite{Najafabadi2015} although they worked successfully in classification problem in many areas. The shallow methods need a good feature representation and assume all data available. Feature engineering focused on constructing features as an essential element of machine learning. However, when the data is rapidly growing within dynamic assumptions such as CD, the handy-crafted feature engineering is very difficult and the traditional methods need to be modified. The popular approach combines Deep Learning method as unsupervised feature representation learner with any supervised classifier from traditional machine learning methods.


In this paper, we proposed a new adaptive scheme of integration between CNN  \cite{LeCun:1,Zeiler2014,Jiuxiang:1} and ELM \cite{huang2006extreme,Huang:ELMSurvey,Huang201532} to handle concept drift in online big stream data. We named it Adaptive CNN-ELM (ACNNELM). We studied ACNNELM scheme for concept drift either changes in the number of feature inputs named virtual drift (VD) or the number of  classes named real drift (RD) or consecutive drift when VD and RD occurred at the same time named hybrid drift (HD) \cite{Budiman_hindawi} in recurrent context (all concepts occur alternately).

We developed ACNNELM based on our previous work on  adaptive ELM scheme named Adaptive OS-ELM (AOS-ELM) that works as single ELM classifier for CD handling \cite{Budiman_hindawi,6922113}. As single classifier, AOS-ELM combines simultaneously many strategies to solve many types of CD in simple platform. 

As our main contributions in this research area, we proposed two models of adaptive hybrid CNN and ELM as follows.
\begin{enumerate}
    \item ACNNELM-1: the adaptive scheme for integrating CNN and ELM to handle concept drift in classifier level;
\item ACNNELM-2: the concatenation aggregation ensemble of integrating CNN and ELM classifiers to boost the performance and to handle concept drift for adaptivity in ensemble level.
\end{enumerate}

Section 1 gives introduction and research objectives. We describe related works in Section 2. Section 3 describes our proposed methods. We focus on the empirical experiments to prove the methods in MNIST and not-MNIST image classification task in Section 4. Section 5 discusses conclusions, challenges, and future directions. 

\subsection{Notations}
We used the notations  throughout this paper to make easier for the readers:

\begin{itemize}
\item Matrix is written in uppercase bold (i.e., $\mathbf{X}$). 
\item Vector is written in lowercase bold (i.e., $\mathbf{x}$).
\item The transpose of a matrix $\mathbf{X}$ is written as $\mathbf{X}^{^\mathrm{T}}$. The pseudo-inverse of  a matrix $\mathbf{H}$ is written as $\mathbf{H}^{\dagger}$. 
\item $\textstyle{g}$ will be used as non linear activation function, i.e., sigmoid, reLU or tanh function. 
\item The amount of training data is $\textstyle{N}$. Each input data $\mathbf{x}$ contains some $\textstyle{d}$ attributes. The target has $\textstyle{m}$ number of classes.
An input matrix $\mathbf{X}$ can be denoted as $\mathbf{X}_{\textstyle{d}\times\textstyle{N}}$ and the target matrix $\mathbf{T}$ as $\mathbf{T}_{\textstyle{N}\times\textstyle{m}}$.
\item We denote the subscript font with parenthesis to show the time sequence number.  The $\mathbf{X_{(0)}}$ is the subset of input data at time $k=0$ as the initialization stage. $\mathbf{X_{(1)}}$,$\mathbf{X_{(2)}}$,...,$\mathbf{X_{(k)}}$ are the subset of input data at the next sequential time. Each subset may have different number of quantity. The corresponding label data is presented as  $\mathbf{T} = \left[\mathbf{T_{(0)}},\mathbf{T_{(1)}},\mathbf{T_{(2)}},...,\mathbf{T_{(k)}} \right]$. 
\item We denote the subscript font without parenthesis to show the concept number. $\textstyle{S}$ concepts (sources or contexts) is using the symbol $\mathbf{X_{s}}$ for training data and $\mathbf{T_{s}}$ for target data. 
\item We denote the drift event  using the symbol $\genfrac{}{}{0pt}{}{\ggg}{VD}$, where the subscript font shows the drift type. I.e., the Concept 1 has virtual drift event and  replaced by Concept 2 (Sudden changes) symbolized as $\mathbf{C}_1 \genfrac{}{}{0pt}{}{\ggg}{VD} \mathbf{C}_2$. The Concept 1 has real drift event and replaced by Concept 1 and Concept 2 recurrently  (Recurrent context) in the shuffled composition symbolized as $\mathbf{C}_1 \genfrac{}{}{0pt}{}{\ggg}{RD} shuffled(\mathbf{C}_1,\mathbf{C}_2)$. 


\end{itemize}

\section{Literature Reviews}

\subsection{Extreme Learning Machine (ELM)}
Extreme Learning Machine (ELM) works based on generalized pseudoinverse for non iterative learning in single hidden layer feedforward neural network (SLFN) architecture  \cite{huang2006extreme,Huang:ELMSurvey,Huang201532}.  Compared with Neural Networks (NN) including CNN, ELM and NN used random value parameters. However, ELM used set and fixed random value in hidden nodes parameters and used non iterative generalized pseudoinverse optimization process, while NN used iterative gradient descent optimization process to smooth the weight parameters.   


The ELM learning objective is to get Output weight ($\beta$), where

\begin{equation}
\label{eq:2}
\hat{\beta} = \textbf{H}^{\dagger}\textbf{T}
\end{equation}

\noindent which $\textbf{H}^{\dagger }$ is Pseudoinverse (Moore-Penrose generalized inverse) of $\textbf{H}$.
The ELM learning is simply equivalent to find the smallest least-squares solution for $\hat{\beta}$ of the linear system $\textbf{H}\beta = \textbf{T}$ when the Output weight $\hat{\beta}$ = $\textbf{H}^{\dagger}\textbf{T}$. 

Hidden layer matrix $\textbf{H}$ is computed using activation function $\textit{g}$ of the summation matrix  from the hidden nodes parameter (input weight $a$ and bias  $b$) with training input \textbf{x} with size N number of training data and L number of hidden nodes $g(a_{i}\cdot\textbf{x}+b_{i})$ (feature mapping). 

To solve $\textbf{H}^{\dagger}$, ELM  used orthogonal projection  based on  ridge regression method, where a positive  1/$\lambda$ value  is added as regularization to the auto correlation matrices  $\textbf{H}^{T}\textbf{H}$ or $\textbf{H}\textbf{H}^{T}$. We can solve Eq. \ref{eq:2} as follows. 

\begin{equation}
	\label{eq:orthogonalization}
	\beta = \left( \frac{\textbf{I}}{\lambda}+\textbf{H}^{T}\textbf{H}\right)^{-1}\textbf{H}^{T}\textbf{T}
\end{equation}


ELM has capability to learn online stream data (data comes one by one or by block with fixed or varying quantity) by solving $\textbf{H}^{T}\textbf{H}$ or $\textbf{H}\textbf{H}^{T}$ by following two methods:


\begin{enumerate}
    \item Sequential series using block matrices inverse.
    
    A Fast and Accurate Online Sequential named online sequential extreme learning machine (OS-ELM) \cite{4012031} has  sequential learning phase. In sequential learning, the $\beta_{(k)}$ computed  as the previous $\beta_{(k-1)}$ function.
    If we have $\hat{\mathbf{\beta}}_{(0)}$ from  $\mathbf{H}_{(0)}$ filled by the $N_0$ as initial training data and $\mathbf{H}_{(1)}$ filled by $N_1$ incremental training data, then the output weights $\hat{\mathbf{\beta}}_{(1)}$ are approximated by solving:
    \begin{equation}
    	\label{eq:oselm}
    \Big ( \left[ \begin{array}{c} \mathbf{H}_{(0)} \\ \mathbf{H}_{(1)} \end{array} \right]^{^\mathrm{T}} \left[ \begin{array}{c} \mathbf{H}_{(0)} \\ \mathbf{H}_{(1)} \end{array} \right] \Big )^{-1} \left[ \begin{array}{c} \mathbf{H}_{(0)} \\ \mathbf{H}_{(1)} \end{array} \right]^{^\mathrm{T}} \left[ \begin{array}{c} \mathbf{T}_{(0)} \\ \mathbf{T}_{(1)} \end{array} \right]
    \end{equation}

    The OS-ELM assumes no changes in the hidden nodes number. However, increasing the hidden nodes number may improve the performance named Constructive Enhancement OS-ELM \cite{4597855}. 
    
    \item Parallelization using MapReduce framework.
    
    Another approach is Elastic Extreme Learning Machine (E$^{2}$LM) \cite{Xin2015464}  or Parallel ELM \cite{He:2013} based on MapReduce framework to solve large sequential training data  in a parallel way.
    First, Map is the transform process of intermediate matrix multiplications for each training data portion. Second, Reduce is the aggregate process to sum the Map result. 

    If $\mathbf{U}= \mathbf{H}^{T}\mathbf{H}$ and $\mathbf{V}= \mathbf{H}^{T}\mathbf{T}$, we have decomposable matrices in $k$ time sequences and can be written as : 
    
    \begin{equation}
	\label{eq:decomU}
	\mathbf{U} = \sum_{k=0}^{k=\infty} \mathbf{U}_{(k)}
    \end{equation}

    \begin{equation}
	\label{eq:decomV}
	\mathbf{V} = \sum_{k=0}^{k=\infty} \mathbf{V}_{(k)}
    \end{equation}

Finally, The corresponding output weights $\beta$ can be obtained with centralized computing using result from reduce/aggregate process. Therefore, E$^{2}$LM learning is efficient for rapidly massive training data set \cite{Xin2015464}. 

    \begin{equation}
	\label{eq:elasticELM}
	\beta = \left( \frac{\textbf{I}}{\lambda}+\textbf{U}\right)^{-1}\textbf{V}
    \end{equation}

\end{enumerate}


E$^{2}$LM is more computing efficient, better performance and support parallel computation than OS-ELM \cite{Xin2015464}, but E$^{2}$LM did not address the possibility for hidden nodes increased during the training. OS-ELM, CEOS-ELM,  E$^{2}$LM, and Parallel ELM method did not address concept drift issues; i.e., when the number of  attributes $d$  in $\mathbf{X}_{\textstyle{d}\times\textstyle{N}}$ or the number of  classes  $m$ in $\mathbf{T}_{\textstyle{N}\times\textstyle{m}}$  in data set has changed. We categorized OS-ELM,  CEOS-ELM, E$^{2}$LM , and Parallel ELM as non-adaptive sequential ELM. 

In this paper, we developed parallelization framework to integrate with CNN to solve concept drift issues in sequential learning as enhancement from our previous work Adaptive OS-ELM (AOS-ELM) \cite{Budiman_hindawi}. 

\subsection{Convolutional Neural Networks (CNN)}


Different with SLFN in ELM, a CNN consists of some convolution and sub-sampling layers in feed forward architecture. CNN is the first successful deep architecture that keep the characteristics of the traditional NN. CNN has excellent performance for spatial visual classification \cite{Simard:2003}. The key benefit of CNN comparing with another deep learning methods are using fewer parameters \cite{Zeiler2014}. 



\begin{figure}[!h]
\includegraphics[width=5in]{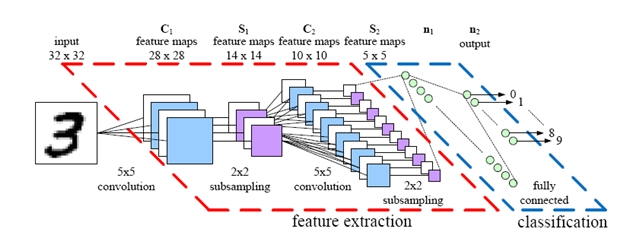}
\caption{CNN Architecture from AlexNet \cite{Zeiler2014}}. 
\label{fig:cnnarch}
\end{figure}

At the end of CNN layer, it is followed by fully connected standard multilayer neural network (See Fig. \ref{fig:cnnarch}) \cite{Zeiler2014}. Many variants of CNN architectures in the literature, but the basic common building blocks are convolutional layer, pooling layer and fully connected layer \cite{Jiuxiang:1}.

The CNN input layer is  designed to exploit the 2D structure with $d \times d \times r$ of image where $d$ is the height and width of  image, and $r$ is the number of channels, (i.e. gray scale image has r=1 and RGB image has r=3). The convolutional layer has $c$ filters (or kernels) of size $k \times k \times q$  where $k < d$  and $q$ can either be the same as the number of channels $r$ or smaller and may vary for each kernel. The filters have locally connected structure which is each convolved with the image to produce $c$ feature maps of size $d - k+1$ (Convolution operations) . 

Each map from convolutional layer is then pooled using either down sampling, mean or max sampling over $s \times s \times s$ contiguous regions ($s$ ranges between 2 for small and up to 5 for larger inputs). An additive bias and activation function (i.e. sigmoid, tanh, or reLU) can be applied to each feature map either before or after the pooling layer. At the end of the CNN layer, there may be any densely connected NN layers for supervised learning  \cite{Zeiler2014}. The learning errors are propagated back to the previous layers using optimization method to finally update the kernel weight parameters and bias.


The convolution operations are heavy computation but  inherently parallel, which  getting beneficial from a hardware parallel implementation \cite{Scherer2010}. Krizhevsky \textit{et. al.}  showed a large, deep CNN is capable of achieving record breaking results on a highly challenging dataset (the 1.2 million high-resolution images in the ImageNet LSVRC-2010 contest into the 1000 different classes) using purely supervised learning. However, the CNN network size is still limited mainly by the amount of memory available on current GPUs \cite{Krizhevsky_imagenetclassification}.


\subsection{CNN ELM Integration}

Huang \textit{et. al.} \cite{bai2015} explained the ELM theories are not only valid for fully connected ELM architecture but are also actually valid for local connections, named local receptive fields (LRF) or similar with Kernel in CNN term. Huang \textit{et. al.} proposed ELM-LRF that has connections between input layer and hidden nodes are randomly generated following any continuous different types of probability distributions. According to Huang \textit{et. al.} the random convolutional hidden nodes CNN is one type of local receptive fields. Different with CNN, ELM with LRF type hidden nodes  keeps the essence of ELM for non iterative output weights calculation. 

Pang \textit{et. al.} \cite{ShanPang_hindawi} implemented deep convolutional ELM (DC-ELM). It uses multiple convolution layers and pooling layers for high level features abstraction from input images. Then, the abstracted features are classified by an ELM classifier. Pang \textit{et. al.} did not use sequential learning approach. Their results give 98.43\% for DC-ELM compared with 97.79\% for ELM-LRF on MNIST regular 15K samples training data. 

Guo \textit{et. al.} \cite{cnn:guo} introduced an integration model of CNN-ELM, and applied to handwritten digit recognition. Guo \textit{et. al.} used CNN as an automatic feature extractor and ELM to replace the original classification layer of CNN. CNN-ELM achieved an error rate of 0.67\% that is lower than CNN and ELM alone. Guo \textit{et. al.} trained the original CNN until converged. The last layer of CNN was replaced by ELM to complete classification without iteration. The experiments used regular MNIST data set with size-normalized and centered in a fixed-size image 28$\times$ 28 pixels. According to Guo \textit{et. al.} numbers filters of different convolution layer have a significant influence on the generalization ability.


\begin{figure}[!h]
\includegraphics[width=5in]{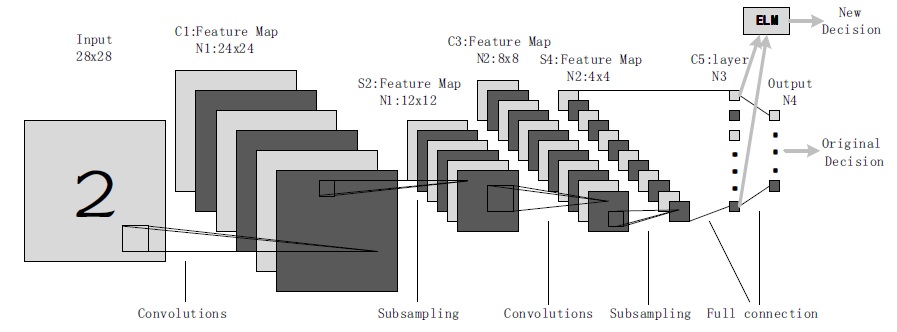}
\caption{Structure of the deep hybrid CNN-ELM model \cite{cnn:guo}}. 
\label{fig:cnnelmhybrid}
\end{figure}



Lee \textit{et. al.} \cite{cnn:lee2} proposed an integration of CNN with OS-ELM. A learning method called an orthogonal bipolar vector (OBV) was also well applied by analyzing the neural networks learned by combining both algorithms. The experimental results demonstrate that the proposed method can conduct network learning at a faster rate than conventional CNN. In addition, it can be used to solve the local minima and overfitting problems. The experiment used NORB dataset,  MNIST handwritten digit standard dataset, and  CIFAR-10 dataset. The recognition rate for MNIST testing data set is 93.54\% .

\subsection{Concept Drift}

The brief explanation of concept drift (CD) has been well introduced by Gama, \textit{et. al.} \cite{Gama:2014:SCD:2597757.2523813} and Minku \cite{5156502} based on Bayesian decision theory for class output $c$ and incoming data $X$.


The concept is the whole distribution (joint distribution P(\textbf{X},c) in a certain time step.
The CD represents any changes in the joint distribution when  $P(c|\textbf{X})$ has changed; i.e., $\exists\textbf{X}:P_{(0)}(\textbf{X},c) \neq P_{(1)}(\textbf{X},c) $, where $P_{(0)}$ and  $P_{(1)}$ are respectively the joint distribution at time $k_{(0)}$ and $k_{(1)}$. 
The CD type has categorization as follows \cite{Gama:2014:SCD:2597757.2523813,2192-6352,5975223,1647649}.
\begin{enumerate}
\item Real Drift (RD) refers to changes in  $P(c|\textbf{X})$. 
The change in  $P(c|\textbf{X})$ may be caused by a change in the class boundary (the number of classes) or the class conditional probabilities (likelihood) $P(\textbf{X}|c)$. 

\item Virtual Drift (VD) refers to the changes in the distribution of the incoming data (i.e., $P(\textbf{X})$ changes). These changes may be due to incomplete or partial feature representation of the current data distribution. The trained model is built with additional data from the same environment without overlapping the actual class boundaries. 
\end{enumerate}

We introduced the third category named Hybrid Drift (HD) when the RD and VD occurred consecutively. \cite{Budiman_hindawi}. 

\begin{figure}[!h]
\includegraphics[width=3.5in]{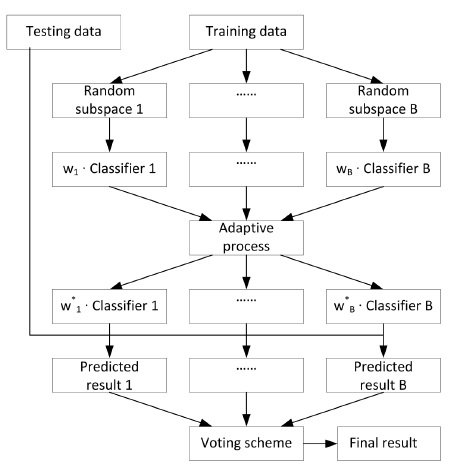}
\caption{Overview of singly adaptive ensemble learning for the random subspace-based classifier ensemble approach. \cite{yu:1}. }
\label{fig:adaptive_ensemble}
\end{figure}

Ensemble learning is the common approaches to tackle concept drift, in which are combined using a form of voting \cite{Tsymbal:2008:DIC:1297420.1297577,39313598}. The ensemble approach can integrate the results of individual classifiers into a unified predicted result to improve the accuracy and robustness than single classifiers \cite{Zang2014}. Yu , \textit{et. al.} \cite{yu:1} proposed a general hybrid adaptive ensemble learning framework  (See Fig. \ref{fig:adaptive_ensemble}). Liu, \textit{et. al.} \cite{liu:1} proposed an ensemble based ELM (EN-ELM)  which uses the cross-validation scheme to build ELM classifiers ensemble. 

Each drift employed different solution strategies. The solution for RD is entirely different from VD. Certain CD requirement needs to replace entirely the outdated concept (concept replacement) either sudden or gradually, but another requirement needs to handle old and new concepts that come alternately (recurring concept). Thus, it is hard to combine simultaneously many types and requirement of complex drift solutions, such as hybrid drift in a simple ensemble platform. The adaptive ensemble approach may be not practical and flexible if each member itself is not designed to be adaptive \cite{Budiman_hindawi} and may need to recall the previous training data (not single pass learning  \cite{978-3-540-22144-9}). Moreover, another simple approach is using  single classifier \cite{MirzaL16,grachten:1,Budiman_hindawi}.

Mirza \textit{et. al.} \cite{MirzaL16} proposed OS-ELM for imbalanced and concept drift tackling named meta-cognitive OS-ELM (MOS-ELM) that was developed based on Weighted OS-ELM (WOS-ELM) \cite{Zong:2013}. MOS-ELM used an additional weighting matrix to control the CD adaptivity, however, it works for RD with concept replacement only. 

Our previous AOS-ELM works as single unified solution for VD, RD, and HD. Also, it can be applied for concept replacement and recurring \cite{Budiman_hindawi} by using simple matrix adjustment and multiplication. We explained AOS-ELM  for each scenario as follows.

\begin{enumerate}
\item Scenario 1: Virtual Drift (VD).

According to interpolation theory from ELM point of view and Learning Principle I of ELM Theory \cite{s12559-014-9255-2}, the input weight and bias as hidden nodes $\mathbf{H}$ parameters are independent of training samples and their learning environment through randomization. Their independence is not only in initial training stage but also in any sequential training stages.  Thus, we can adjust the input weight and bias pair $\{\mathbf{a}_i, \mathbf{b}_i\}^{L}_{i=1} $ on any sequential  stages and still have probability one that $\|\mathbf{H}\mathbf{\beta} - \mathbf{T} \|< \epsilon$ to handle additional feature inputs.

\begin{figure}[!h]
		\centerline{\includegraphics[width=4in]{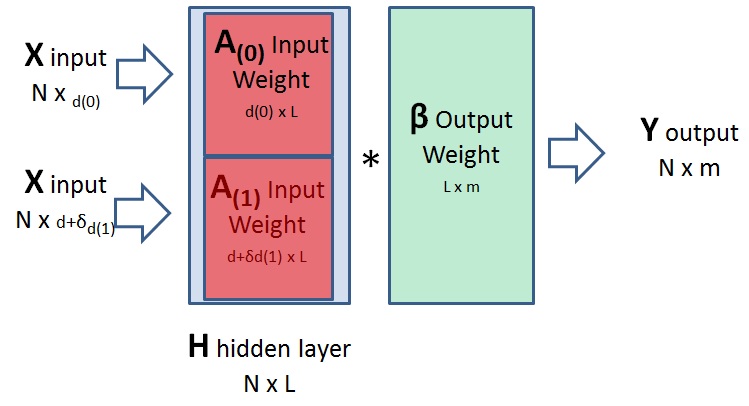}}
	\caption{The AOS-ELM Virtual Drift Tackling  }
	\label{fig:aos-vd}
\end{figure}

\item Scenario 2: Real Drift (RD).

According to universal approximation theory and inspired by the related works \cite{4597855}, the AOS-ELM has real drift capability by modifying the output matrix with zero block matrix concatenation to change the size matrix dimension without changing the norm value. Zero block matrix means the previous $\mathbf{\beta}_{(k-1)}$ has no knowledge about the new concept. ELM can approximate to any complex decision boundary,  as long as the output weights ($\mathbf{\beta}_(k)$) are kept minimum when the number of output classes increased. 

\begin{figure}[!h]
		\centerline{\includegraphics[width=4in]{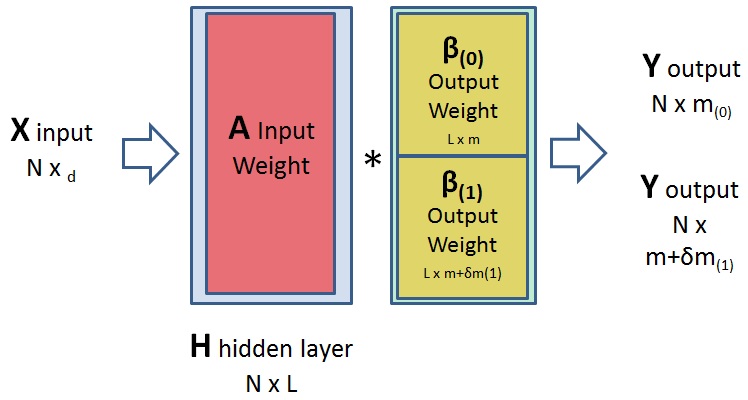}}
	\caption{The AOS-ELM Real Drift Tackling  }
	\label{fig:aos-rd}
\end{figure}

\item Scenario 3: Hybrid drift (HD).

Hybrid drift is a consecutive drifts scenario when the VD and RD occur in the same drift event. Scenario 1 (VD) requires modification on hidden nodes \textbf{H} parameters and scenario 2 (RD) requires modification on output weight $\beta$. Both modifications are independence, thus, they can be modified in the same event. The example implementation is when we need to combine from different type of training data set. 

\begin{figure}[!h]
		\centerline{\includegraphics[width=4in]{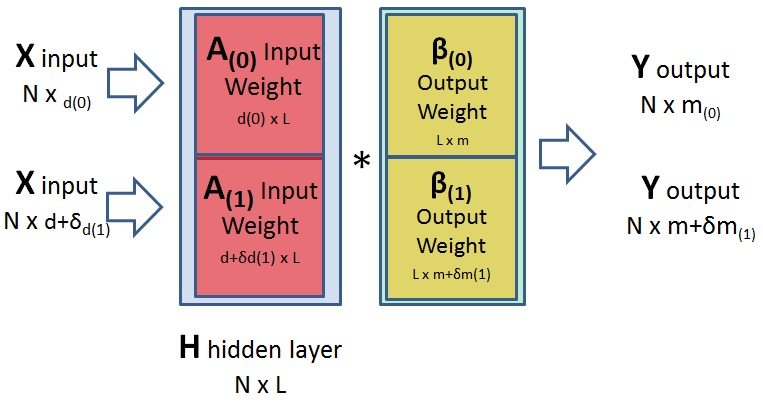}}
	\caption{The AOS-ELM Hybrid Drift Tackling  }
	\label{fig:aos-hd}
\end{figure}
 
\end{enumerate}

\subsection{CNN in Concept Drift }

Grachten \textit{et. al.} \cite{grachten:1} proposed some adapting strategies for single CNN classifier system based on two common adaptive approaches: 1) REUSE is to reuse the model upon the change of task and replace the old task training data with the new task training data. 2) RESET is to ignore the previous representations learned, and begin the new task learning with a randomly initialized model. Further, Grachten \textit{et. al.} categorized as follows. 
\begin{enumerate}
\item RESET: Initialize parameters with random values;
\item RESET PRF: Combination between the prior regularization on convolutional filters (PRF) with the RESET option.  PRF improves a bit the RESET baseline sometimes, and the gains are usually moderate;
\item REUSE ALL:  Initialize all parameters from prior model (except output layer);
\item REUSE CF: Selectively reuse the prior model by keeping previous convolutional filters (CF) from the prior model. 
\end{enumerate}

\begin{figure}[!h]
\includegraphics[width=4in]{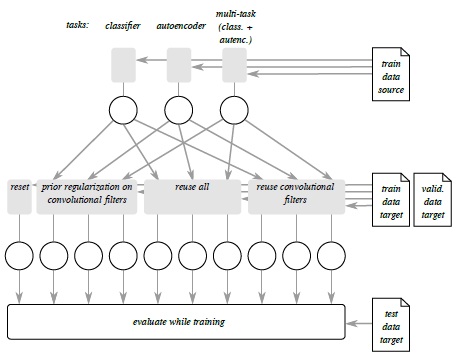}
\caption{Schematic overview of Grachten \textit{et. al.}'s experiment for a single data set; training methods in grey rounded boxes, represent models in circles,  data instances in document shapes, and the evaluation method in white rounded box \cite{grachten:1}}. 
\label{fig:grachten}
\end{figure}

Grachten \textit{et. al.} divided regular MNIST into two subsets (Data Set 1 from class 0 to 4, and Data Set 2 from class 5 to 9). The training set is  50000 data, 10000 data for validation and 10000 testing data. The classifier used a CNN with a convolutional layer (32 feature maps using kernels size $5 \times 5$, sigmoid activations and Dropout), followed by a max-pooling layer ($2 \times 2$ pool size). The classification stage used a fully connected layer (40 sigmoid units and 10 softmax units). 

Zhang \textit{et. al.} \cite{Zhang2016} proposed an adaptive CNN (ACNN), whose structure automatic expansion based on the average system error and training recognition rate performance. The incremental learning for new training data is handled by new branches addition while the original network keep unchanged. ACNN used the structure global expansion until meet the average error criteria and local expansion to expand the network structure. Zhang \textit{et. al.} used ORL face database. The model has recognition rate increased from 91.67\% to 93.33\% using local expansion. However, Zhang \textit{et. al.} did not discuss any concept drift handling. 

 

According to Zhang \textit{et. al.} no such theory about how CNN structure constructed, i.e. the number of layers, the number of feature maps per layer. Researchers constructed and  compared each CNN candidate performance for the best one.  Some studies tried to use hardware acceleration to speed up the performance comparison discovery. However, Zhang \textit{et. al.} did not discuss any concept drift handling. We used the idea of global expansion in our proposed method. 

\begin{figure}[!h]
\includegraphics[width=4in]{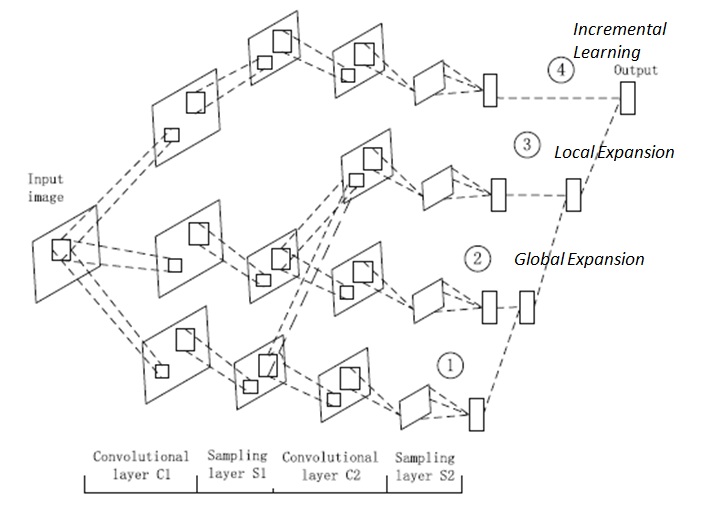}
\caption{The Adaptive CNN architecture with Global expansion (2), Local  expansion (3) and Incremental Learning (4) \cite{Zhang2016}}. 
\label{fig:Zhang2016}
\end{figure}

\section{Proposed Method}

We used common CNN-ELM integration \cite{ShanPang_hindawi,bai2015,cnn:guo,cnn:lee2} architecture   when the last convolution layer output is fed as hidden nodes weight \textbf{H} of ELM (See Fig. \ref{fig:cnnelmhybrid}). However, for final \textbf{H}, we used nonlinear  optimal tanh ($1.7159\times tanh (\frac{2}{3}\times\mathbf{H}$) activation function \cite{LeCun:1998} to have better generalization accuracy. We used also CNN global expansion structure \cite{Zhang2016} to improve the performance accuracy. 

\begin{figure}[!h]
\includegraphics[width=3in]{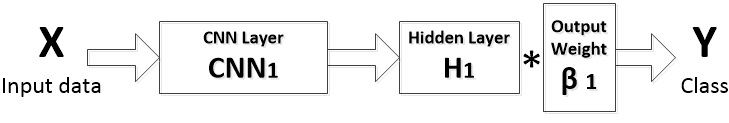}
\caption{CNN-ELM integration architecture : The last convolution layer output is submitted as hidden nodes weight \textbf{H} of ELM}. 
\label{fig:method2}
\end{figure}

We used the E$^2$LM as a parallel supervised classifier to replace fully connected NN. Compared with regular ELM method, we do not need input weight as hidden nodes parameter (See Fig. \ref{fig:method2}). 

We deployed the integration architecture becomes two models: 1) ACNNELM-1 : This ACNNELM model works based on AOS-ELM for concept drift adaptability on classifier level. 2) ACNNELM-2 : This model combines some ACNNELMs and aggregated as matrices concatenation ensemble to handle concept drift adaptability on ensemble level.

\begin{figure}[!h]
\includegraphics[width=4in]{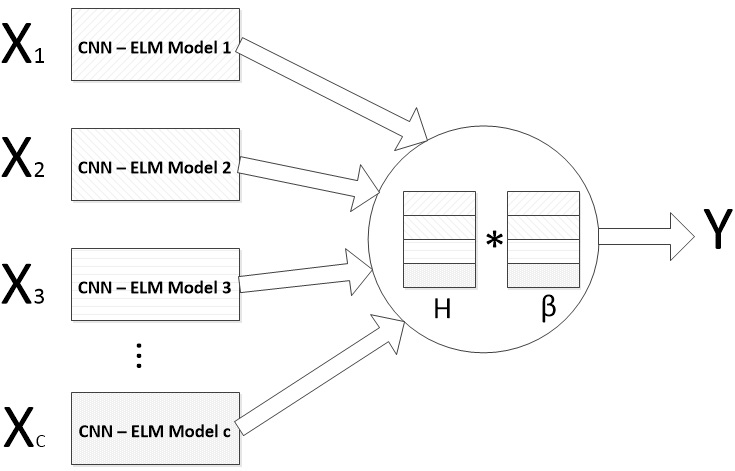}
\caption{The aggregation concatenation ensemble of hybrid CNN ELM method  to boost the performance and to handle concept drift for adaptivity in ensemble level (ACNNELM-2)} 
\label{fig:method_combine}
\end{figure}

In ACNNELM-1, we used matrices $\mathbf{U}$ and $\mathbf{V}$ adjustment padded by using zero block matrices before recomputing the $\beta$ (See Equation \ref{eq:elasticELM}). In ACNNELM-2, we enhanced the model by matrices concatenation aggregation of  \textbf{H} and $\beta$ as the result from CNN ELM individual model (See Fig. \ref{fig:method_combine}). 

\begin{figure}[!h]
\includegraphics[width=4in]{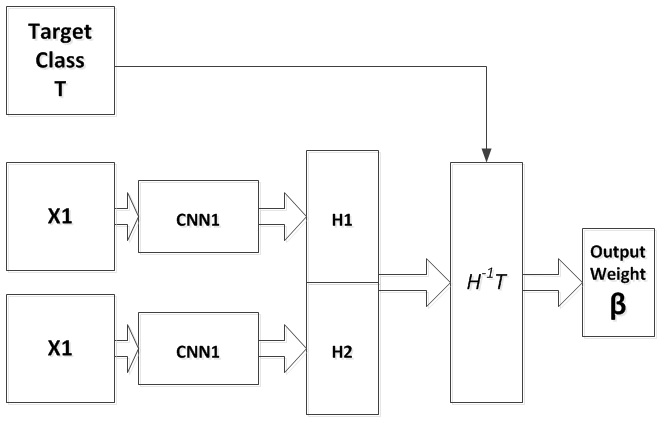}
\caption{Adaptive scheme for hybrid CNN and ELM  for concept drift handling in classifier level (ACNNELM-1)}. 
\label{fig:method7}
\end{figure}

We developed each method  for some handling scenarios as belows. 
\begin{enumerate}
    \item ACNNELM-1
    \begin{enumerate}
        \item Virtual drift (VD).
        
        Let's $\mathbf{X_2}$ is additional features from new concept, we assigned the $\mathbf{X_2}$ to new CNN model then concatenated together the last layer of all CNN models to be single \textbf{H} matrix to compute $\beta$ (See Fig. \ref{fig:method7}). The new $\mathbf{H}_(k)$ has larger column size than previous $\mathbf{H}_(k-1)$. Because we used E$^2$LM method, the $\mathbf{H}_(k)$ change needs  $\mathbf{U}$ and $\mathbf{V}$ to be adjusted also ($\mathbf{U}= \mathbf{H}^{T}\mathbf{H}$ (auto correlation matrix) and $\mathbf{V}= \mathbf{H}^{T}\mathbf{T}$).
        
         We need to adjust the previous $\mathbf{U}_{(k-1)}$ by padding zero block matrix in row and column to have same row and column with $\mathbf{U}_{(k)}$ to compute:
        
        \begin{equation}
	\label{eq:ACNNELM-1U}
	\sum_{0}^{k} \mathbf{U} = \sum_{0}^{k-1} \mathbf{U}_{(k-1)} + \mathbf{U}_{(k)}
    \end{equation}
    
        We need to adjust the previous $\mathbf{V}_{(k-1)}$ by padding zero block matrix in row only to have same row  with $\mathbf{V}_{(k)}$ to compute:
        
        \begin{equation}
	\label{eq:ACNNELM-1V}
	\sum_{0}^{k} \mathbf{V} = \sum_{0}^{k-1} \mathbf{V}_{(k-1)} + \mathbf{V}_{(k)}
    \end{equation}
        
        \item Real Drift (RD).
        
        Let's $\mathbf{T_2}$ is additional output classes expansion as a new concept. Thus we need to modify the ELM layer only by adjusting the column dimension of matrix $\mathbf{V}$ with zero block matrix for column adjustment only. 
        
        \item Hybrid Drift (HD).
        
        The HD solution basically is a combination of VD and RD method. We introduced new CNN model and adjusting the row and column dimension of matrix $\mathbf{V}$.

    \end{enumerate}
    
    \item ACNNELM-2

\begin{enumerate}

    \item Matrices Concatenation Ensemble to boost the performance.
    
    The idea is based on matrix multiplication $\mathbf{H}\beta$ that actually can be decomposable. We used to re-compose ELM members to be one ensemble that seems as one big ELM. This idea only needs minimum two members and no need to have additional multi-classifier strategies (See Fig. \ref{fig:method3}).
    
    \begin{figure}[!h]
    \includegraphics[width=4in]{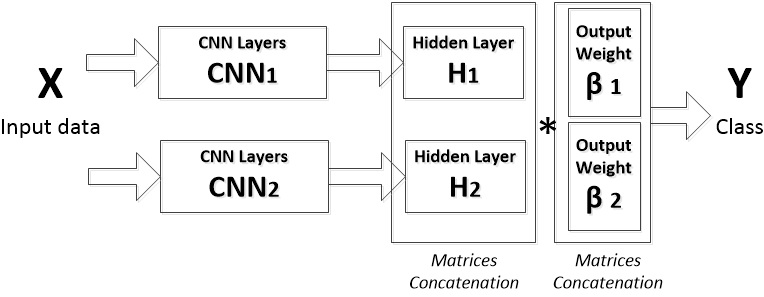}
    \caption{Matrices Concatenation Ensemble}. 
    \label{fig:method3}
    \end{figure}
    
    \item Virtual drift (VD).
    
    We enhanced the concatenation concept for VD handling. Let's $\mathbf{X_2}$ is additional features from new concept, we assigned the $\mathbf{X_2}$ to new CNN-ELM hybrid model, then concatenated the learning result  without disturbing to the old CNN-ELM hybrid model. Using this simple model, we have flexibilities to use reuse or reset strategy \cite{grachten:1} for recurrent or sudden concept drift handling.  
    
    \begin{figure}[!h]
    \includegraphics[width=4in]{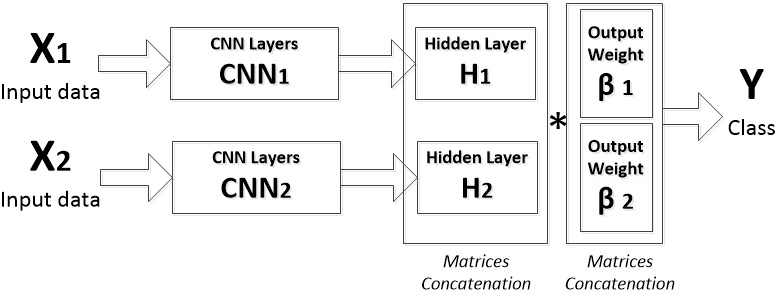}
    \caption{Virtual drift}. 
    \label{fig:method4}
    \end{figure}
    
    \item Real Drift (RD)
    
    We applied the concatenation concept for RD handling also, but in RD case, we need to modify the previous $\beta_1$ concatenated with zero block matrix $\mathbf{0}$ first, so that the total class number is equal with the $\beta_2$. The $\beta_2$ needs training data modification by incrementing the order of training class number. 
    
    \begin{figure}[!h]
    \includegraphics[width=4in]{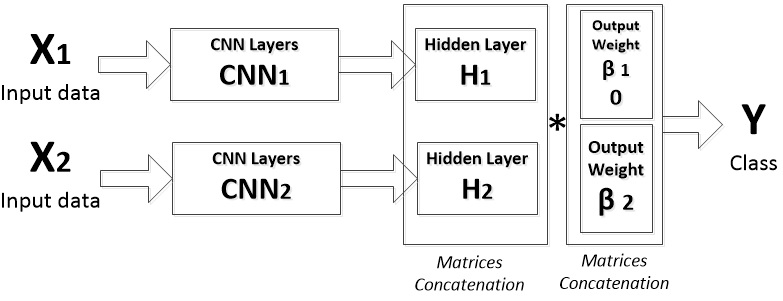}
    \caption{Real Drift}. 
    \label{fig:method6}
    \end{figure}

    \item Hybrid Drift (HD)
    
    VD and RD can be processed on separated CNN-ELM hybrid model parallel. Thus the HD scenario is easy with Matrices Concatenation Ensemble method. 
    
    \begin{figure}[!h]
    \includegraphics[width=4in]{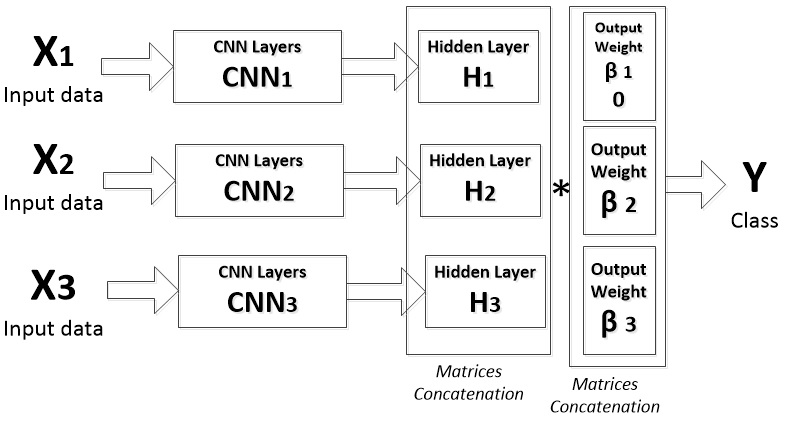}
    \caption{Hybrid Drift}. 
    \label{fig:method5}
    \end{figure}
   
   \cite{notMNIST1} 
    
\end{enumerate}
\end{enumerate}

\section{Experiment and Performance Results}

\subsection{Data set}

\begin{figure}[!h]
\centering
\includegraphics[width=2in]{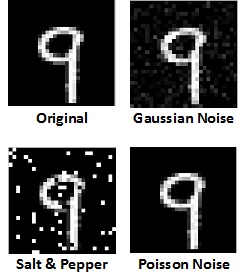}
\caption{Extended MNIST Data set by adding random gaussian, salt\&pepper, poisson noise to original data. }. 
\label{fig:noise}
\end{figure}

Dataset is the successful key for this research to simulate big stream data. MNIST is the common data set for big data machine learning, in fact, it accepted as standard and give an excellent result. MNIST data set is a balanced data set that contains numeric handwriting (10 target class) with size $28 \times 28$ pixel in a gray scale image. The dataset has been divided for 60,000 examples for training data and separated 10,000 examples for testing data  \cite{lecun-mnisthandwrittendigit-2010}. However, to simulate big stream data, the regular MNIST is not adequate. For that reason, we developed extended MNIST data set with larger training examples by adding 3 types of image noises (See Fig. \ref{fig:noise}). Our extended MNIST data set finally has 240,000 examples of training data and 40,000 examples of testing data.

We also enhanced the $28 \times 28$ image size as attributes with additional attributes based on Histogram of oriented gradient (HOG) of images with size $9 \times 9$. The total attributes become 865 attributes. The HOG additional attributes have been used in our research \cite{Budiman_hindawi} to simulate VD scenario. 

\begin{figure}[!h]
\includegraphics[width=2in]{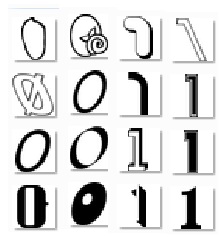}
\caption{Not-MNIST Data set for numeric symbols }. 
\label{fig:nonMNIST1}
\end{figure}

\begin{figure}[!h]
\includegraphics[width=2in]{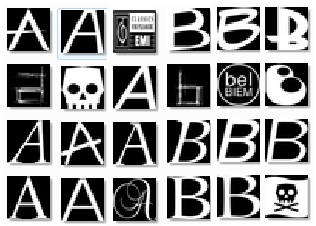}
\caption{Not-MNIST Data set for alphabet A-J symbols }. 
\label{fig:nonMNIST}
\end{figure}

We used not-MNIST large data set for additional experiments. We expect not-MNIST is a harder task than MNIST. Not-MNIST dataset has a lot of foolish images (See Fig. \ref{fig:nonMNIST1} and \ref{fig:nonMNIST}). Bulatov explained the logistic regression on top of stacked autoencoder with fine-tuning gets about 89\% accuracy whereas the same approach gives got 98\% on MNIST \cite{notMNIST1}. Not-MNIST has gray scale $28 \times 28$ image size as attributes. We divided the set to be numeric (360,000 data) and alphabet (A-J) symbol (540,000) data including many foolish images. The challenge with not-MNIST numeric and not-MNIST alphabet is many similarities between class 1 with class I, class 4 with class A, and another similar foolish images. 

\subsection{Experiment Methods}

We defined the scope of works as following:
\begin{enumerate}
 
\item We enhanced DeepLearn Toolbox \cite{DLtoolbox} with Matlab parallel computing toolbox. 
\item We used single precision than double precision for all computation in this paper. 
Double precision has accuracy improvement in matrix inverse computation than single precision. The ELM in Matlab using double precision by default and it has better accuracy (i.e. 4-5\% higher) than single precision. However, using single precision has more CPU and memory saving especially for parallelization in big stream data. Unfortunately, most papers did not mention how their precision method used. We want to achieve  accuracy improvement not because of precision computation factor.
\item We focused on simple CNN architecture that consist of convolution layers (c), following by \textit{reLU} activation layer  then sample pooling layer (s) in this paper. 
\item We used extended MNIST data set and not-MNIST data set to simulate big stream data.

\end{enumerate} 

We summarized the experiment methods in the following tables \ref{alltable1} and \ref{alltable2}.
\begin{table}[!h]

\renewcommand{\arraystretch}{1.3}
    \caption{Concept Drift Scenarios, Model Architectures, and Computing Resources}
    \label{alltable1}
    \centering
       \subfloat[Concept Drift Sequential Patterns\label{table:flow}]{%
\centering     
\begin{tabular}{|p{1.25cm}|p{1.5cm}|p{2.25cm}|}
\hline
Data Set & Sequential Patterns Scenarios  & Cause of shift\\
\hline
\hline
MNIST &  Sudden change , Recurring Context & Additional attributes or classes  \\
\hline
Not MNIST &  Sudden change , Recurring Context & Additional classes  \\
\hline
\end{tabular}}

\subfloat[Model Architectures\label{table:arc}]{%
\centering     
\begin{tabular}{|p{2cm}|p{5cm}|}
\hline
Model & Architecture  \\
\hline
\hline
Non Adaptive OS-ELM & Robust ELM, sig, Regularization $C=10^3$, L=1000, 2500, 3000, 5000, Batch=1000, single   \\
\hline
CNN &  relu, 4 Layers with 6c-2s-12c-2s, Kernel size=5, pooling=down, Batch=60000, single   \\
\hline
ACNNELM-1 & relu, 2 Layers with 3c-2s, Kernel size=5, pooling=down, Batch=60000, Regularization $C=10^3$, single    \\
\hline
ACNNELM-2 & relu, 4 Layers with 6c-2s-12c-2s, Kernel size=5, pooling=down, Batch=60000, Regularization $C=10^3$, single     \\
\hline
ACNNELM-2 & relu, 4 Layers with 12c-2s-18c-2s, Kernel size=5, pooling=down, Batch=60000, Regularization $C=10^5$, single     \\
\hline
\end{tabular}}
  \end{table}

\begin{table}[!h]

\renewcommand{\arraystretch}{1.3}
    \caption{Data set Dimension, Quantity, Evaluation method, and Performance Measurement.}
    \label{alltable2}
    \centering
    \subfloat[Data Set dimension and  Quantity\label{table:dataset}]{%
\centering     
\begin{tabular}{|c|c|c|c|}
\hline
Data Set Concepts &  Inputs & Outputs & Data   \\
\hline
\hline
$\mathbf{MNIST}_1$ & 784  & 10 (0-9)    & 240,000   \\
\hline
$\mathbf{MNIST}_2$ & 865  & 10 (0-9)    & 240,000   \\
\hline
$\mathbf{MNIST}_3$& 784  & 6 (0-5)     & 150,000   \\
\hline
$\mathbf{MNIST}_4$ & 784  & 4 (6-9)    & 100,000   \\
\hline
$\mathbf{Not MNIST}_1$& 784  & 10 (0-9)     & 360,000   \\
\hline
$\mathbf{Not MNIST}_2$ & 784  & 10 (A-J)    & 540,000   \\
\hline
$\mathbf{Not MNIST}_3$ & 865  & 10 (0-9)    & 360,000   \\
\hline
$\mathbf{Not MNIST}_4$ & 865  & 20 (A-J,0-9)    & 900,000   \\
\hline
\end{tabular}
    }
    \\
    \subfloat[Evaluation  Method\label{table:eval}]{%
\centering     
\begin{tabular}{|p{1.25cm}|p{2cm}|c|c|}
\hline
Data Set & Evaluation Method & Training & Testing \\
\hline
\hline
MNIST &  Holdout ($5\times$ trials on different computers)   & 240,000 & 40,000 \\
\hline
Not-MNIST &  Cross Validation 5 Fold   & 720,000 & 180,000 \\
\hline
\end{tabular}
}
    \\
    \subfloat[Performance Measurements\label{table:mea}]{%
\centering     
\begin{tabular}{|p{2cm}|p{4cm}|}
\hline
Measure & Specification \\
\hline
\hline
Accuracy &  The accuracy of classification in \% from $\frac{\#\textit{Correctly Classified}}{\#\textit{Total Instances}} $ \\
\hline
Testing Accuracy &  The accuracy measurement of the testing data which not part of training set. \\
\hline
Cohen's Kappa and kappa error  &  The  statistic measurement of inter-rater agreement for categorical items. \\
\hline
\end{tabular}
    }
  \end{table}

To verify our method, we designed some experiments to answer the following research questions: 
\begin{itemize}

\item How is the performance comparison of more CNN layer added?
\item How is the performance comparison of non linear  \textit{optimal tanh} function compared with another function? 
\item How is the performance comparison between non adaptive OS-ELM, CNN, and our method CNN-ELM (ACNNELM-1 and ACNNELM-2) using the extended MNIST  and not-MNIST data set?
\item How is the effectiveness of Matrices Concatenation Ensemble of ACNNELM-2 method improved the performance if we used up to 16 independent models? 
\item How does the ACNNELM-1 and ACNNELM-2 handle VD, RD and HD using simulated scenario on extended MNIST data set and not-MNIST data set? 

In VD scenario, the drift event is :

$\mathbf{MNIST}_1 \genfrac{}{}{0pt}{}{\ggg}{VD} \mathbf{MNIST}_2$

$\mathbf{Not-MNIST}_1 \genfrac{}{}{0pt}{}{\ggg}{VD} \mathbf{Not-MNIST}_3$

In RD scenario, the drift event is :

$\mathbf{MNIST}_3 \genfrac{}{}{0pt}{}{\ggg}{RD} shuffled(\mathbf{MNIST}_{3,4})$

$\mathbf{Not-MNIST}_1 \genfrac{}{}{0pt}{}{\ggg}{RD} shuffled(\mathbf{Not MNIST}_{1,2})$



In HD scenario, the drift event is :
 

$\mathbf{MNIST}_3 \genfrac{}{}{0pt}{}{\ggg}{HD} \mathbf{MNIST}_2$

$\mathbf{Not-MNIST}_1 \genfrac{}{}{0pt}{}{\ggg}{HD} \mathbf{Not-MNIST}_4$
  
\end{itemize}

\subsection{Performance Results}

\begin{figure}[!h]
\includegraphics[width=5in]{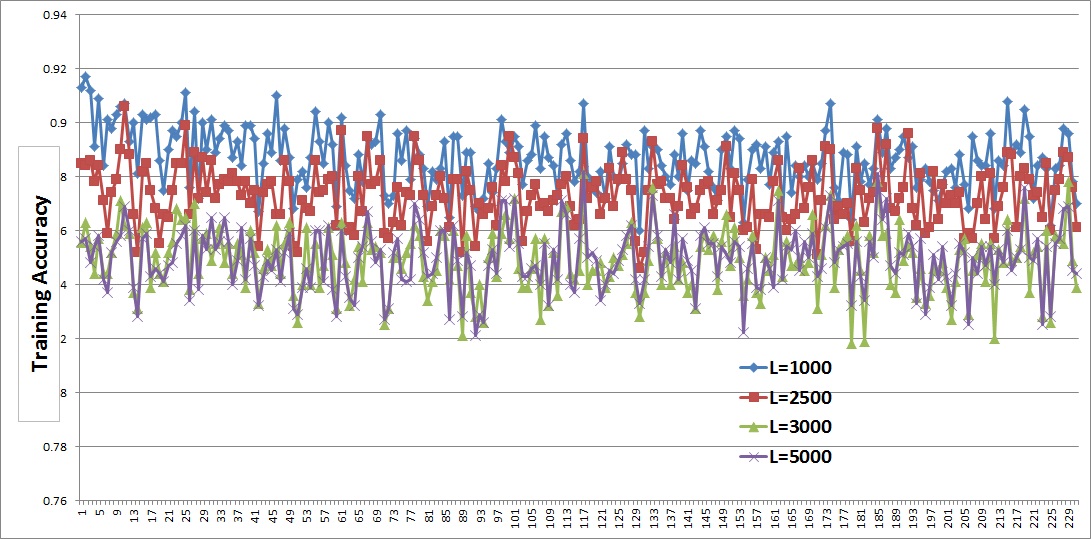}
\caption{ELM Result on extended MNIST dataset for L=1000, L=2500, L=3000, L=5000 with C=0.5. }. 
\label{fig:elm_result}
\end{figure}

For benchmark, we compared the accuracy performance with Non Adaptive OS-ELM. We can not guarantee the performance for big sequential training data unless we find the best hidden nodes parameter and regularization scalar and using double precision (See Fig. \ref{fig:elm_result}). In our extended MNIST, the testing accuracy using single precision experiment is  for L=1000 is 88.73\%, L=2500 is 87.88\%, L=3000 is 85.48\%, L=5000 is 86.21\% . In our Not-MNIST Numeric, the testing accuracy using single precision experiment is  for L=1000 is 79.02\%, L=2000 is 77.48\%, L=4000 is 76.50\%, L=6000 is 74.31\% . No guarantee for larger hidden nodes gives better accuracy performance using ELM in single precision. 

For concept drift capability, we expect no accuracy performance decreased after the drift event, no matter how big training data is required compared with its full batch version. 

\begin{figure}[!h]
\includegraphics[width=5in]{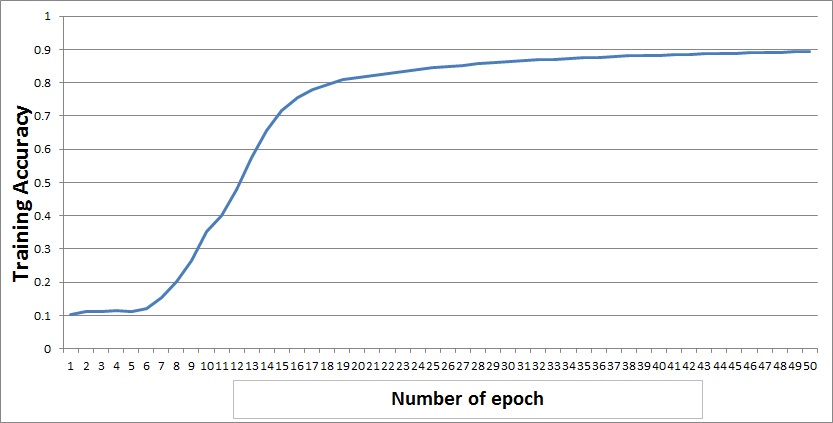}
\caption{CNN Result on extended MNIST dataset 4 Layers with 6c-2s-12c-2s }. 
\label{fig:cnn_result}
\end{figure}

Different with CNN, the performance of CNN for big sequential training data is better using a larger number of the epoch. CNN has better scalability than ELM for big sequential learning data, but it needs longer time for iteration to improve the performance (See Fig. \ref{fig:cnn_result}). With epoch=50, the testing accuracy is 90.32\%. From the learning time perspective, the time for 50 iterations is equivalent to build 50 models of individual CNN ELM  sequentially.  

We compared above benchmark result with our CNN-ELM method: 

\begin{enumerate}

        \item The performance of CNN-ELM can be improved by using more layers in expanded structure. In Table \ref{result_table1}, the model 6c-2s-12c-2s has better accuracy than model 6c-2s. 
        
        \begin{table}[!h]

\renewcommand{\arraystretch}{1.3}
    \caption{The effectiveness of CNN layer number  for CNN-ELM performance improvement. The experiment was in 5 $\times$ trial using different model in extended MNIST. }
    \label{result_table1}
    \centering
    \begin{tabular}{|c|c|c|}
\hline
Layer &  Testing Accuracy \% & Cohen Kappa \% \\
\hline
\hline
6c-2s &  91.32$\pm$0.52    & 90.36 (0.16)   \\
\hline
6c-2s-12c-2s & 94.29$\pm$0.79    & 93.65 (0.13)   \\
\hline
\end{tabular}
    
\end{table} 
        
        \item The performance of CNN-ELM can be improved by using \textit{optimal tanh} activation function (See table \ref{result_function})
        
\begin{table}[!h]

\renewcommand{\arraystretch}{1.3}
    \caption{The effectiveness of non linear activation function for \textbf{H} in performance improvement. The experiment was in 5 $\times$ trial using different function in extended MNIST with  1 6c-2s ACNNELM-1 model and 3 6c-2s ACNNELM-2 model.}
    \label{result_function}
    \centering
    \begin{tabular}{|c|c|c|c|}
\hline
Model &  Function & Testing Accuracy \% & Cohen Kappa \% \\
\hline
\hline
ACNNELM-1 & No  & 91.32$\pm$0.52    & 90.36 (0.16)   \\
\hline
ACNNELM-1 & $Sigmoid$  & 78.58$\pm$1.27    & 75.29 (1.61)    \\
\hline
ACNNELM-1 & $Softmax$   & 90.38$\pm$0.97     & 89.25 (0.84)    \\
\hline
ACNNELM-1 & $tanh_opt$ & 91.46$\pm$0.37    & 90.52 (0.43)   \\
\hline
ACNNELM-2 & No  & 91.81$\pm$0.15    & 90.12 (0.23)   \\
\hline
ACNNELM-2 & $tanh_opt$  & 93.52$\pm$0.18    & 92.49 (0.17)   \\
\hline

\end{tabular}
    
\end{table}




    \item The effectiveness of Matrices Concatenation Ensemble of ACNNELM-2 up to  16 Models that generated asynchronously.
    
    In this experiment, we studied that concatenation ensemble concept has better performance than standalone model  (See Figure \ref{fig:concatenation} and \ref{fig:concatenation_nonMNIST}). It can retain the performance well even the standalone model performance seems decreased. By using 6c-2s-12c-2s, we only have 192 nodes per model that used as hidden nodes of ELM. Multiplicated by 16 models, we have 3072 ELM hidden nodes, compared with 5000 hidden nodes of regular ELM.  

    \begin{figure}[!h]
\centering
\includegraphics[width=3in]{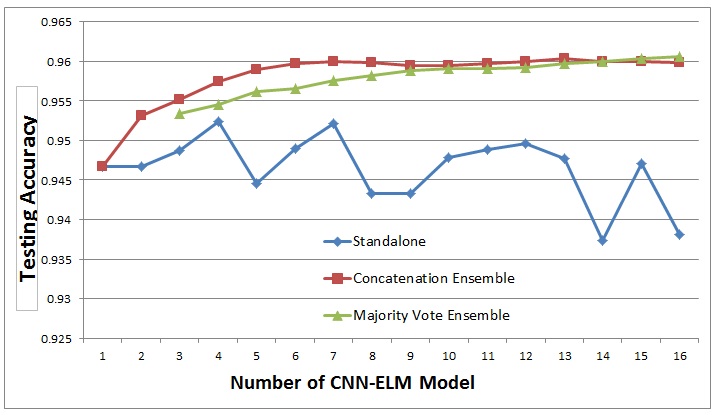}
\caption{Testing Accuracy Comparison between standalone model vs ACNNELM-2 concatenation ensemble  vs majority vote ensemble up to 16 models with 6c-2s-12c-2s in extended MNIST }. 
\label{fig:concatenation}
\end{figure}

\begin{figure}[!h]
\centering
\includegraphics[width=3in]{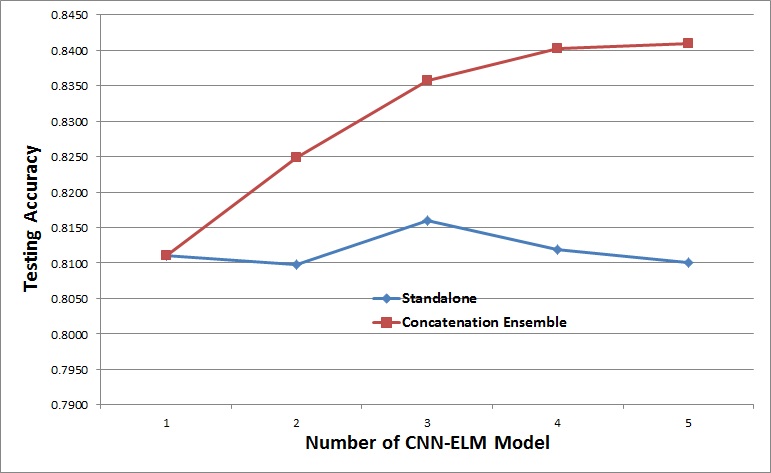}
\caption{Testing Accuracy Comparison between standalone model vs ACNNELM-2 concatenation ensemble  up to 5 models with 6c-2s-12c-2s in Not MNIST Numeric }. 
\label{fig:concatenation_nonMNIST}
\end{figure}

    \item Virtual drift handling
    
    \begin{enumerate}
        \item ACNNELM-1
        
         In this experiment, we verified our VD handling method in ACNNELM-1. First, we used a single model and trained with $\mathbf{MNIST}_1$ concept. For the same model, we trained with the next  $\mathbf{MNIST}_2$ concept. We build the new CNN 6c-1s for additional attributes. The last layers of all CNN then combined to modify the ELM layer. 

\begin{table}[!h]

\renewcommand{\arraystretch}{1.3}
    \caption{Virtual drift handling. The ACNNELM-1 used Model 1 6c-2s and Model 2  6c-1s model for MNIST and Model 1 12c-2s-18c-2s and Model 2 12c-1s for Not MNIST Numeric}
    \label{result_tablevd}
    \centering
    \begin{tabular}{|c|c|p{1.5cm}|p{1.5cm}|}
\hline
Model &  Concept  & Testing Accuracy \% & Cohen Kappa \% \\
\hline
\hline
Model 1 & $\mathbf{MNIST}_1$  & 91.46$\pm0.37$    & 90.14 (0.16)   \\
\hline
Model 2 & $\mathbf{MNIST}_2$  & 94.54$\pm0.20$    & 93.10 (0.26)   \\
\hline
Model 1 & $\mathbf{Not-MNIST}_1$  & 81.17$\pm0.46$    & 80.14 (0.54)   \\
\hline
Model 2 & $\mathbf{Not-MNIST}_3$  &
82.95$\pm0.42$     & 81.06 (0.16)   \\
\hline
\end{tabular}  
\end{table}

        \item ACNNELM-2
        
        In this experiment, we verified our VD handling method in ACNNELM-2. First, we picked up the previous CNN-ELM models (Model 1 and Model 2) in $\mathbf{MNIST}_1$ concept. Then, for $\mathbf{MNIST}_2$ concept, we build the new CNN-ELM models (Model 3 and Model 4) but using the additional attributes only (81 attributes or 9 $\times$ 9 size). The Model 3 and Model 4 have 6c-1s-12c-1s, kernel size=3, 300 Hidden nodes to ELM . We concatenated all models to be one concatenation ensemble and compared the current and previous performance (Table \ref{result_tablevd1}). The CNN-ELM model has no dependencies with another model (No shared parameters).  We tested all models with 10 classes testing data set.
        
    
    \begin{table}[!h]

\renewcommand{\arraystretch}{1.3}
    \caption{Virtual drift handling in ACNNELM-2 for extended MNIST. Model 1,2 6c-2s and Model 3,4 6c-1s}
    \label{result_tablevd1}
    \centering
    \begin{tabular}{|c|c|p{1.5cm}|p{1.5cm}|}
\hline
Model &  Concept  & Testing Accuracy \% & Cohen Kappa \% \\
\hline
\hline
Model 1 & $\mathbf{MNIST}_1$  & 93.19$\pm0.24$     & 92.58 (0.26)   \\
\hline
Model 2 & $\mathbf{MNIST}_1$  & 91.65$\pm0.56$   & 90.45 (0.22)    \\
\hline
Model 1|2 & $\mathbf{MNIST}_1$  & 93.77$\pm0.28$     & 92.78 (0.25)    \\
\hline
Model 1|2|3 & $\mathbf{MNIST}_2$ & 95.29$\pm0.17$   & 93.55 (0.13)   \\
\hline
Model 1|2|3|4 & $\mathbf{MNIST}_2$  & 95.57$\pm0.73$    & 94.01 (0.14)   \\
\hline

\end{tabular}  
\end{table}

   \begin{table}[!h]

\renewcommand{\arraystretch}{1.3}
    \caption{Virtual drift handling in ACNNELM-2 for Not-MNIST Numeric. Model 1,2 12c-2s-18c-2s and Model 3 12c-1s}
    \label{result_tablevd3}
    \centering
    \begin{tabular}{|c|c|p{1.5cm}|p{1.5cm}|}
\hline
Model &  Concept  & Testing Accuracy \% & Cohen Kappa \% \\
\hline
\hline
Model 1 & $\mathbf{Not-MNIST}_1$  & 81.17$\pm0.46$    & 80.14 (0.54)   \\
\hline
Model 1|2 & $\mathbf{Not-MNIST}_1$  & 83.47$\pm0.27$    & 81.28 (0.29)   \\
\hline
Model 1|2|3 & $\mathbf{Not-MNIST}_3$ & 86.34$\pm0.59$    & 84.38 (0.27)   \\
\hline
\end{tabular}  
\end{table}
        
    \end{enumerate}

    \item Real drift handling.
    
    \begin{enumerate}
        \item ACNNELM-1
        
        In this experiment, we verified our RD handling method in ACNNELM-1. First, we used single model trained with $\mathbf{MNIST}_3$ concept. For the same model, we continued with the next  $\mathbf{MNIST}_4$ and  $\mathbf{MNIST}_3$ concept without building any new CNN model. We repeat the experiment on Not-MNIST data set. 
        
        \begin{table}[!h]

\renewcommand{\arraystretch}{1.3}
    \caption{Real  drift handling. The ACNNELM-1 used Model 1  6c-2s model for extended MNIST and Model 1 12c-2s-18c-2s for Not-MNIST }
    \label{result_tablerd1}
    \centering
    \begin{tabular}{|c|p{2cm}|p{1.5cm}|p{1.5cm}|}
\hline
Model &  Concept  & Testing Accuracy \% & Cohen Kappa \% \\
\hline
\hline
Model 1 & $\mathbf{MNIST}_3$  & 58.46$\pm0.53$    & 53.27 (0.17)   \\
\hline
Model 1 & $\mathbf{MNIST}_{3,4}$  & 92.45 $\pm0.63$     & 91.12 (0.25)    \\
\hline
Model 1 & $\mathbf{Not-MNIST}_1$  & 34.46$\pm1.23$    & 30.42 (1.17)   \\
\hline
Model 1 & $\mathbf{Not-MNIST}_{1,2}$  & 79.12 $\pm0.58$     & 77.22 (0.45)    \\
\hline

\end{tabular}  
\end{table}
        
        \item ACNNELM-2
    
    In this experiment, we verified our RD handling method in ACNNELM-2. First, we build the  ACNNELM-2 models (Model 5 and Model 6)  in $\mathbf{MNIST}_3$ concept with only 6 classes (with order number 1 to 6). Then, for $\mathbf{MNIST}_4$ concept, we build the new CNN-ELM models (Model 7 and Model 8) using 10 classes (continuing the class order number from 7 to 10). We concatenated all models to be one concatenation ensemble and compared the current and previous performance (Table \ref{result_tablerd2}). In the matrices concatenation, we need to adjust the $\beta$ of Model 5 and Model 6 (only have 6 columns) by using additional zero block matrix to pad the columns to be 10 columns. Then we can concatenate with Model 7 and Model 8. We tested all models with 10 classes testing data set. 
    
        \begin{table}[!h]

\renewcommand{\arraystretch}{1.3}
    \caption{Real  drift handling. The ACNNELM-2 used 6c-2s-12c-2s model for extended MNIST}
    \label{result_tablerd2}
    \centering
    \begin{tabular}{|c|c|p{1.5cm}|p{1.5cm}|}
\hline
Model &  Concept  & Testing Accuracy \% & Cohen Kappa \% \\
\hline
\hline
Model 5 & $\mathbf{MNIST}_3$  & 58.57$\pm0.12$    & 53.96 (0.27)   \\
\hline
Model 6 & $\mathbf{MNIST}_3$  & 58.16$\pm0.28$   & 54.00 (0.27)    \\
\hline
Model 5|6|7 & $\mathbf{MNIST}_{3,4}$  & 93.53$\pm0.19$     & 92.26 (0.25)    \\
\hline
Model 5|6|7|8 & $\mathbf{MNIST}_{3,4}$ & 93.61$\pm0.52$   & 92.55 (0.23)   \\
\hline

\end{tabular}  
\end{table}

\begin{table}[!h]
\renewcommand{\arraystretch}{1.3}
    \caption{Real  drift handling. The ACNNELM-2 used 12c-2s-18c-2s model for Not-MNIST}
    \label{result_tablerd3}
    \centering
    \begin{tabular}{|c|p{2cm}|p{1.5cm}|p{1.5cm}|}
\hline
Model &  Concept  & Testing Accuracy \% & Cohen Kappa \% \\
\hline
\hline
Model 1 & $\mathbf{Not-MNIST}_1$  & 34.46$\pm1.23$    & 30.42 (1.17)   \\
\hline
Model 1|7 & $\mathbf{Not-MNIST}_{1,2}$  & 81.99$\pm0.47$   & 81.01 (0.10)    \\
\hline

\end{tabular}  
\end{table}

\end{enumerate}
    
    \item Hybrid drift handling
    
    \begin{enumerate}
        \item ACNNELM-1
        
        In this experiment, we verified our HD handling method in ACNNELM-1. First, we used single model trained with $\mathbf{MNIST}_3$ concept. For the same model, we just continued the training with the next  $\mathbf{MNIST}_4$ concept in the same time with building 1 CNN model for additional attributes. 
        
  \begin{table}[!h]

\renewcommand{\arraystretch}{1.3}
    \caption{Hybrid drift handling. The ACNNELM-1 used Model 1 6c-2s and Model 2  6c-1s model for MNIST and Model 1 12c-2s-18c-2s and Model 2 12c-1s for Not MNIST.}
    \label{result_tablehd1}
    \centering
    \begin{tabular}{|c|c|p{1.5cm}|p{1.5cm}|}
\hline
Model &  Concept  & Testing Accuracy \% & Cohen Kappa \% \\
\hline
\hline
Model 1 & $\mathbf{MNIST}_3$  & 58.46$\pm0.53$    & 53.27 (0.17)   \\
\hline
Model 2 & $\mathbf{MNIST}_2$  & 93.42$\pm0.32$    & 91.10 (0.16)   \\
\hline
Model 1 & $\mathbf{Not-MNIST}_1$  & 34.46$\pm1.23$    & 30.42 (1.17)   \\
\hline
Model 2 & $\mathbf{Not-MNIST}_4$  & 84.29$\pm0.42$    & 82.10 (0.36)   \\
\hline
\end{tabular}  
\end{table}              
        
        \item ACNNELM-2
    
    In this experiment, we verified our HD handling method in ACNNELM-2. Simply, We just combine the VD and HD models to be one concatenation ensemble. 
    
     \begin{table}[!h]

\renewcommand{\arraystretch}{1.3}
    \caption{Hybrid  drift handling. The ACNNELM-2 used Model 5-8 6c-2s-12c-2s model and Model 3,4 6c-1s for extended MNIST.}
    \label{result_tablehd2}
    \centering
    \begin{tabular}{|c|c|p{1.5cm}|p{1.5cm}|}
\hline
Model &  Concept  & Testing Accuracy \% & Cohen Kappa \% \\
\hline
\hline
Model 5 & $\mathbf{MNIST}_3$  & 58.57$\pm0.12$    & 53.96 (0.27)   \\
\hline
Model 5|6|7|3 & $\mathbf{MNIST}_2$  &  95.30$\pm0.26$    &  94.78 (0.24)   \\
\hline
Model 5|6|7|8|3|4 & $\mathbf{MNIST}_2$ &  95.94$\pm0.17$  &  95.49 (0.22) \\
\hline

\end{tabular}  
\end{table}

 \begin{table}[!h]

\renewcommand{\arraystretch}{1.3}
    \caption{Hybrid  drift handling. The ACNNELM-2 used Model 1 12c-2s-18c-2s model and Model 3 12c-1s for Not-MNIST.}
    \label{result_tablehd3}
    \centering
    \begin{tabular}{|c|c|p{1.5cm}|p{1.5cm}|}
\hline
Model &  Concept  & Testing Accuracy \% & Cohen Kappa \% \\
\hline
\hline
Model 1 & $\mathbf{Not-MNIST}_1$  & 34.46$\pm1.23$    & 30.42 (1.17)   \\
\hline
Model 1|3 & $\mathbf{Not-MNIST}_4$  &  85.14$\pm0.22$    & 83.35 (0.29)   \\
\hline

\end{tabular}  
\end{table}

\end{enumerate}

\end{enumerate}

\section{Conclusion}

The proposed method gives better adaptive capability for classifier level (ACNNELM-1) and ensemble level (ACNNELM-2).  ACNNELM-2 has better computation scalability and performance accuracy than ACNNELM-1 as result of the aggregation ensemble benefit.

However, some CNN related parameters need to be further investigated, i.e. iterations, random weight for kernel assignment, error backpropagation optimization, decay parameters, and larger layers for larger feature dimension. Also, We need to investigate and implement CNN ELM for non spatial recognition, i.e. posed based human action recognition \cite{6922113}.

We think some ideas for future research:  

\begin{itemize}
\item We will develop the methods on another CNN framework that fully supported to CUDA GPU computing. The purpose is to increase the scale up capability and to speed up the training time in big image data set. 
\item We need to investigate another optimum learning parameters, i.e., stochastic gradient descent, the optimum kernel weight, dropout and dropconnect regularization,  decay parameters. To improve the performance, we believe the optimum CNN parameters also work well for CNNELM. 
\end{itemize}

\section{Acknowledgment}
This work is supported by Higher Education Center of Excellence Research Grant funded Indonesia Ministry of Research and Higher Education Contract No. 1068/UN2.R12/ HKP.05.00/2016

\section{Conflict of Interests}
The authors declare that there is no conflict of interest regarding the publication of this paper.


\bibliography{library}

\bibliographystyle{abbrv}

\end{document}